\documentclass[wcp]{jmlr}
\usepackage{longtable}
\usepackage{lineno}
\usepackage{verbatim}
\usepackage{float}
\makeatletter
\let\Ginclude@graphics\@org@Ginclude@graphics 
\makeatother

\jmlrvolume{222}
\jmlryear{2023}
\jmlrworkshop{ACML 2023}

\title[FedOpenHAR]{FedOpenHAR: Federated Multi-Task Transfer Learning for Sensor-Based Human Activity Recognition}
\author{\Name{Egemen İşgüder} \and \Name{{Ö}zlem {Durmaz İncel}} \\ \Email{\{egemen.isguder, ozlem.durmaz\}@boun.edu.tr} \\ \addr Bogazici University, Istanbul, Turkey}

\begin{document}

\maketitle


\begin{abstract}
Motion sensors integrated into wearable and mobile devices provide valuable information about the device users. Machine learning and, recently, deep learning techniques have been used to characterize sensor data. Mostly, a single task, such as recognition of activities, is targeted, and the data is processed centrally at a server or in a cloud environment. However, the same sensor data can be utilized for multiple tasks and distributed machine-learning techniques can be used without the requirement of the transmission of data to a centre. This paper explores \textit{Federated Transfer Learning in a Multi-Task} manner for both sensor-based human activity recognition and device position identification tasks. The OpenHAR framework is used to train the models, which contains ten smaller datasets. The aim is to obtain model(s) applicable for both tasks in different datasets, which may include only some label types. Multiple experiments are carried in the Flower federated learning environment using the DeepConvLSTM architecture. Results are presented for federated and centralized versions under different parameters and restrictions. By utilizing transfer learning and training a task-specific and personalized federated model, we obtained a similar accuracy with training each client individually and higher accuracy than a fully centralized approach. 
\end{abstract}



\section{Introduction}
Motion sensors available on wearable and mobile devices are commonly used to characterize their users: activities performed, who and where they are, and where they carry their devices. Particularly, monitoring people's physical activity levels can help them stay active and reduce their risk of chronic diseases such as obesity. In the literature, machine learning and, recently, deep learning techniques are applied to the sensor data. Although there are efforts to train and run the learning algorithms on the device, still the wearables are resource-limited regarding computation and memory. Hence, processing is performed at a central server or a cloud environment. Wearable and mobile devices are personal devices, and the integrated sensors collect privacy-sensitive personal data. Centrally, the collection and processing of such data may violate the users' privacy. In this respect, federated learning is an emerging approach. 

Federated learning is a novel technique in machine learning first announced by Google. The main goal of federated learning is to treat data in a privacy-preserving manner. Especially with the recent updates on the General Data Protection Regulation (GDPR), data privacy has become a significant national and international concern. Federated learning offers a critical solution at this point. Instead of sharing data among learning participants (clients), model parameters are shared, and the learning is realized in a distributed manner. This way, data privacy is assured, and the wireless communication costs decrease because only model parameters are shared between participants instead of big chunks of data. This also helps devices with resource constraints like wearable devices, smartphones and IoT devices.

An area where federated learning is utilized is sensor-based human activity recognition (HAR). Data from movement sensors such as accelerometers and gyroscopes are processed with machine learning techniques to monitor a user's activities (walking, sitting, etc.) or special activities such as sports exercises or activities that may include life-threatening situations (e.g. falling).

In this paper, for the recognition of human activities, DeepConvLSTM~\cite{DCL} is applied for centralized and federated learning, both with various methods and settings. For the learning part, the OpenHAR framework~\cite{OpenHAR}, which consists of ten smaller datasets, is used. In total, it contains accelerometer sensor data positioned in various parts of the body collected from 211 participants. Every dataset is owned by a different client in the federated learning experiments. In contrast, two versions exist in centralized learning: one similar to the federated setup, in which each dataset is used for centralized learning separately at each client, and another form where all data is combined in a big pile for centralized learning. Federated experiments are also categorized as one-task, multi-task and multi-task with a layered hierarchy. The tasks to learn during the experiments are the activities of the users and the positions of the devices. In the OpenHAR dataset, the devices were not located in a specific position. To the best of our knowledge, this paper is the first to implement a multi-task classification problem in multiple different datasets with a federated transfer learning method, where all datasets may not contain every type of label.

For the federated learning part, the layered hierarchy for learning consists of pre-trained, common, task-specific and personalized layers inspired by \cite{Multi-Task}, where they focus on multiple image recognition tasks. One of the clients is selected to pre-train the initial weights, which constitutes the pre-trained layer, and then all clients, without any task restriction, participate in one federated loop, which creates the common layer. Afterwards, there are two federated loops (one for each task) to implement the task-specific layers. Ultimately, every client uses its own data for simple training to obtain the personalized layer. 
In the end, a model with an accuracy of 72.4\% is obtained (which is 0.2\% less than the baseline), which works on multiple tasks, 14 different labels, and opens the path to learning new labels and tasks with its layered hierarchy.

The paper continues as follows: Section~\ref{sec:Related} examines the recent works in the area, Section~\ref{sec:Method} explains the used dataset in experiments, the selected classification algorithm and the federated learning architecture. Section~\ref{sec:Experiments} focuses on the results of experiments realized in different conditions with various challenges. Section~\ref{sec:Conclusion} summarizes the results.


\section{Related Work}
\label{sec:Related}

FedHealth~\cite{Fedhealth} suggests a federated learning platform for wearable healthcare devices. The UCI public dataset for smartphones~\cite{anguita2013public} is used for training in the FedHealth study. This dataset is just one of the OpenHAR-integrated datasets used in our study. In FedHealth, machine learning techniques such as KNN and SVM are used instead of deep learning techniques.

In another study, authors ~\cite{portet20} use yet again a dataset from OpenHAR-integrated datasets, which is RealWorldHAR~\cite{sztyler2016body}, in order to train a deep artificial neural network based on CNN. In addition, they compare the success of different parameter aggregation algorithms (FedAvg, FedPer, FedMA). It is reported that the CNN classifier, alongside the FedAvg algorithm, has achieved the highest results. The FedAR study~\cite{presotto}, which focuses on the challenge of finding labelled data in human activity recognition applications and also semi-supervised machine learning in that context, designed experiments using two separate datasets from OpenHAR. Comparing the results of MLP and CNN-based architectures, the authors report that MLP achieves better success. The same authors ~\cite{presottofedclar} also proposed a clustering-based federated learning approach. It is also reported that FedCLAR achieved better results than FedAvg and FedHealth within both datasets.

In the study of ~\cite{Multi-Task}, the writers propose a novel federated learning architecture combining multi-task and federated learning. Their proposed algorithm is composed of four different types of layers: pre-trained, common, task-specific and personalized. During their experiments, they use an artificially generated network consisting of face images. Their dataset's labels are binary, containing only two different types. They achieve close results with their baseline (individually trained) and their proposed method. In our study, instead of dividing one dataset into separate clients, every client uses a whole different dataset, which may or may not contain all the types of labels, which is the crucial difference. Regarding similar works in the field, although federated multi-task learning was utilized for different tasks in the image recognition domain or federated transfer learning in the human activity recognition domain, this paper is the first to implement human activity recognition with a federated multi-task transfer learning architecture. 


\section{Dataset and Federated Multi-Task Learning}
\label{sec:Method}

\subsection{OpenHAR Framework}

OpenHAR~\cite{OpenHAR}, offers a platform that contains ten public datasets. It consolidates the data from different datasets at different sampling rates to 10 Hz. Data is collected through accelerometers placed on various positions of the body. The whole data has 17 different daily activity labels collected from 211 participants and 14 different body positions.

\begin{figure}[t]
\vspace*{-0.5cm}
	\centering
	\includegraphics[width=0.65\columnwidth]{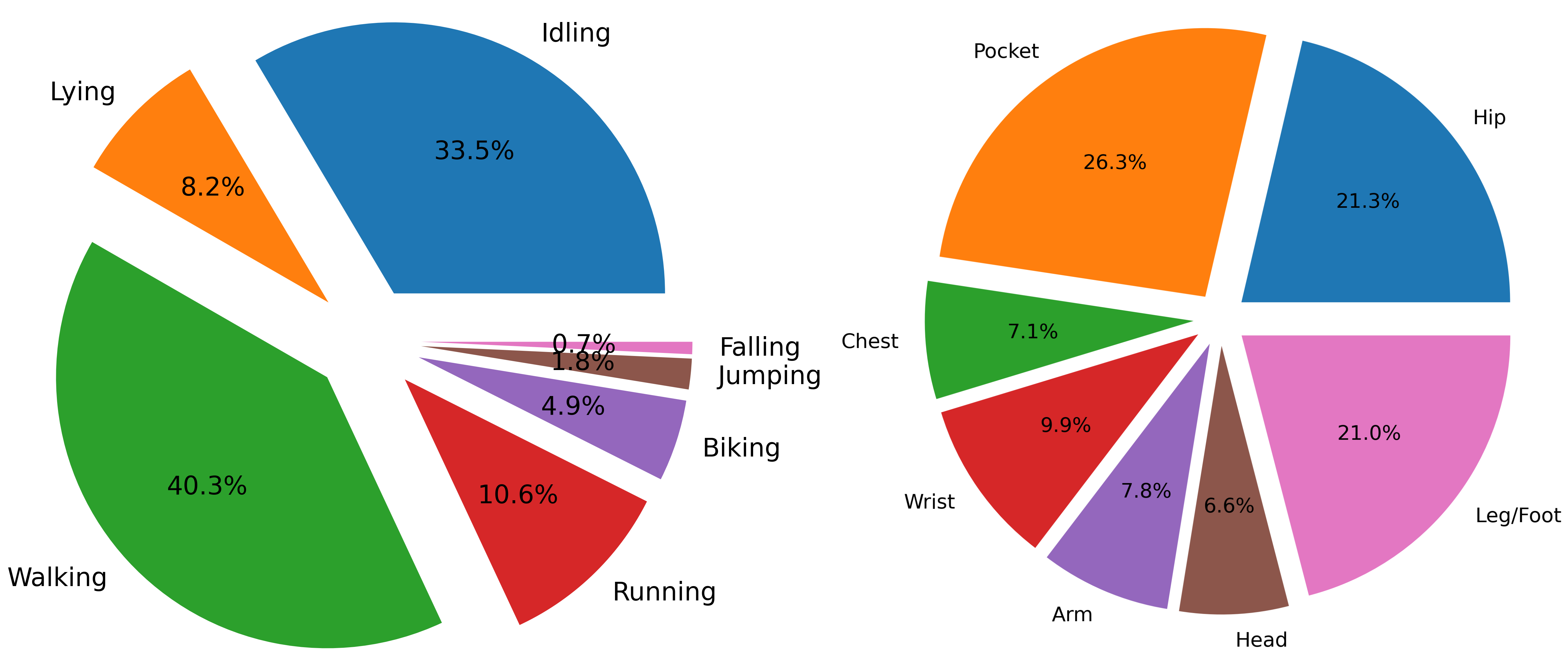}
	\caption{Distribution of Activity (Left) and Position (Right) Labels.}
	\label{fig:Distributions}
\end{figure}

All datasets contain these columns: user ID information, type of activity, sensor position, timestamp and x-y-z axes accelerometer readings. After some initial training experiments, some labels' success rates from the confusion matrix were observed to be less than the others. Since some labels are much scarcer than others, some of the similar labels are combined to obtain a better label distribution. In the data, there were five different labels for ``walking" activity: ``Walking", ``Walking inc. stairs", ``Walking stairs up", ``Walking stairs down", and ``Walking at stairs" which are all combined into one label,``Walking". The same is applied to the position labels, i.e.``Foot, left",``Foot, right" are all combined into``Leg/Foot" label. After preprocessing, the resulting distributions of the labels are given in Figure~\ref{fig:Distributions}.

\subsection{Architecture and Federated Transfer Learning}

We used the DeepConvLSTM architecture ~\cite{DCL} in training the models. It combines Deep Convolutional Neural Network with LSTM (Long Short Term Memory) Network. The architecture of the algorithm consists of four convolutional layers, then two softmax layers and one softmax classifier layer at the end. The convolutional layers of the first part act as feature extractors and provide feature maps of the tabular data. In contrast, the recurrent layers model the temporality of the obtained feature maps. The algorithm is chosen as it is proven to be effective in HAR classification problems.

FedAvg, FedProx, and Federated Matched Averaging are some of the most commonly used averaging algorithms in Federated Learning. This study uses FedAvg as the server's aggregation algorithm \cite{FedAvg}. In FedAvg, the parameter aggregation is based on weighted averages. The central server aggregates all local model parameters into one new global model and then resends the model to a sub-group of clients. The main federated learning algorithm loops until the desired number of communication rounds.

\begin{figure}[H]
\begin{center}
	\centering
	\includegraphics[width=0.51\columnwidth]{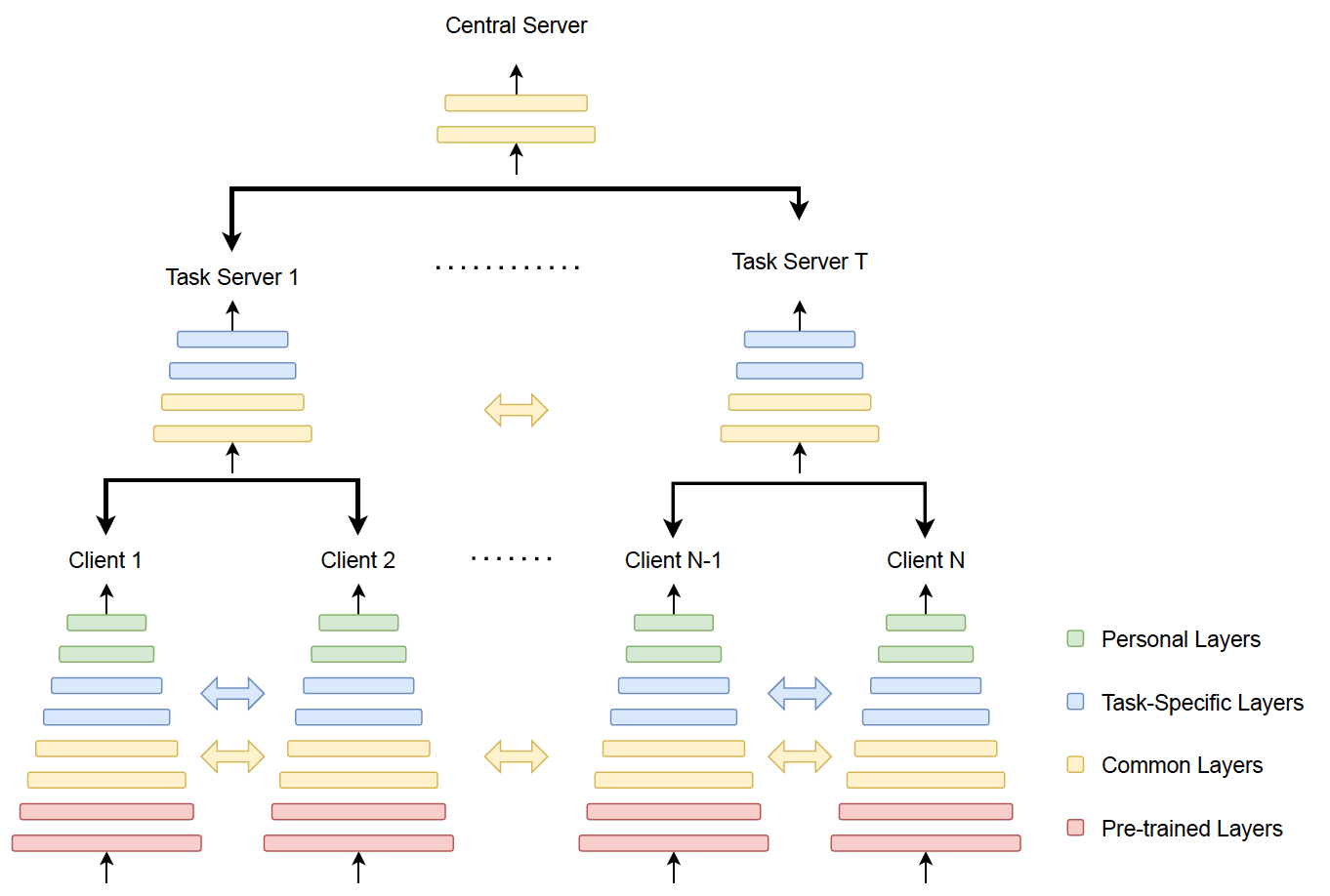}
		\caption{Federated Multi-Task Transfer Learning Architecture (from \cite{Multi-Task}} 
	\label{fig:Hierarchy}
 \end{center}
 
\end{figure}

We use a Federated Multi-Task Transfer learning approach inspired by the architecture in \cite{Multi-Task} for image recognition tasks. The main layered hierarchy can be seen in Figure~\ref{fig:Hierarchy}. However, in this study, instead of using separate servers for each task, the same server trains each layer of the hierarchy, and every time a layer's training is finished, its parameters are frozen, and the training is applied to the subsequent layers.

\vspace*{-1.3cm}
\section{Experiments}
\label{sec:Experiments}

We performed different experiments to evaluate the performance of the explained techniques individually and in combination. For all experiments, data is always split into train and test sets with 80\%-20\% ratio. Flower environment~\cite{Flower} is used in the experiments. 

\subsection{Centralized, Individual and Federated Learning}

In centralized experiments, all data from all datasets are combined into one big pile, which does not conserve data privacy. \textit{Cross entropy loss} is used as the default loss criteria, and \textit{stochastic gradient descent} is used as the optimizer. For the second class of experiments, instead of combining data into one big pile, each client (each dataset) is treated individually, using only their data, conserving data privacy. However, they cannot benefit from other clients' data. The position identification task is only trained with three clients because they are the only clients with more than two types of position labels in their data. The accuracy is reported in terms of the weighted average accuracy of all clients (each client's data size is taken as weight). Individual training is considered as the baseline for further comparison with the federated experiments. Thirdly, in the federated experiments, the setup is similar to the individual training, but this time, there is model parameter sharing, hence the federated learning. For the one-task experiments, only one type of label's clients are trained at a time. The percentages are the weighted average accuracies of each type of the client.

\begin{table}[b!]
\vspace*{-0.4cm}
\centering
\scriptsize
\caption{Accuracies of Centralized, Individual and Federated Experiments.}
\begin{tabular}{|l|c|c|}
\hline
\textbf{Method/Label Type}   & \multicolumn{1}{l|}{\textbf{Activity Label}} & \multicolumn{1}{l|}{\textbf{Position Label}} \\ \hline
\textbf{Centralized in Bulk} & 69.8\%                                       & 59.2\%                                       \\ \hline
\textbf{Individual Training} & 76.7\%                                       & 65.0\%                                       \\ \hline
\textbf{Federated One Task}  & 61.5\%                                       & 57.2\%                                       \\ \hline
\end{tabular}
\label{tab:general_table}
\vspace*{-0.1cm}
\end{table}

The results are presented in Table~\ref{tab:general_table}. Position identification results are lower than the activity recognition results because the number of clients with position data is less than the number of activity clients, and position labels vary more between clients than activity label types. Federated learning results are the lowest because of the heterogeneity between label types in different clients, negatively affecting parameter sharing.

\subsection{Federated Multi-Task and Federated Multi-Task Transfer Learning}

In the multi-task learning experiments, we considered two scenarios. During the simple federated-multi-task experiments, rather than learning one label at a time (separate simulations), both labels are learned in the same experiment. The other method is multi-task transfer learning, which combines federated learning, transfer learning and multi-task learning, as presented in Figure~\ref{fig:Hierarchy}. First, a simple dataset is trained for pre-trained layers. In this case, dataset-9 in OpenHAR (\cite{sztyler2016body}) is used since, for position identification, it revealed the best results. Then, the next layer is trained with all 13 clients, which creates the common layers. Afterwards, two different training sessions were performed for activity (10 clients) and position (3 clients) tasks, which resulted in the task-specific layers. Finally, every client is fine-tuned, and the last layers of their models are trained with their own data. Before training each layer, the previous layers are frozen. 

\begin{figure}[H]
\vspace*{-0.2cm}
	\centering
\includegraphics[width=0.45\columnwidth]{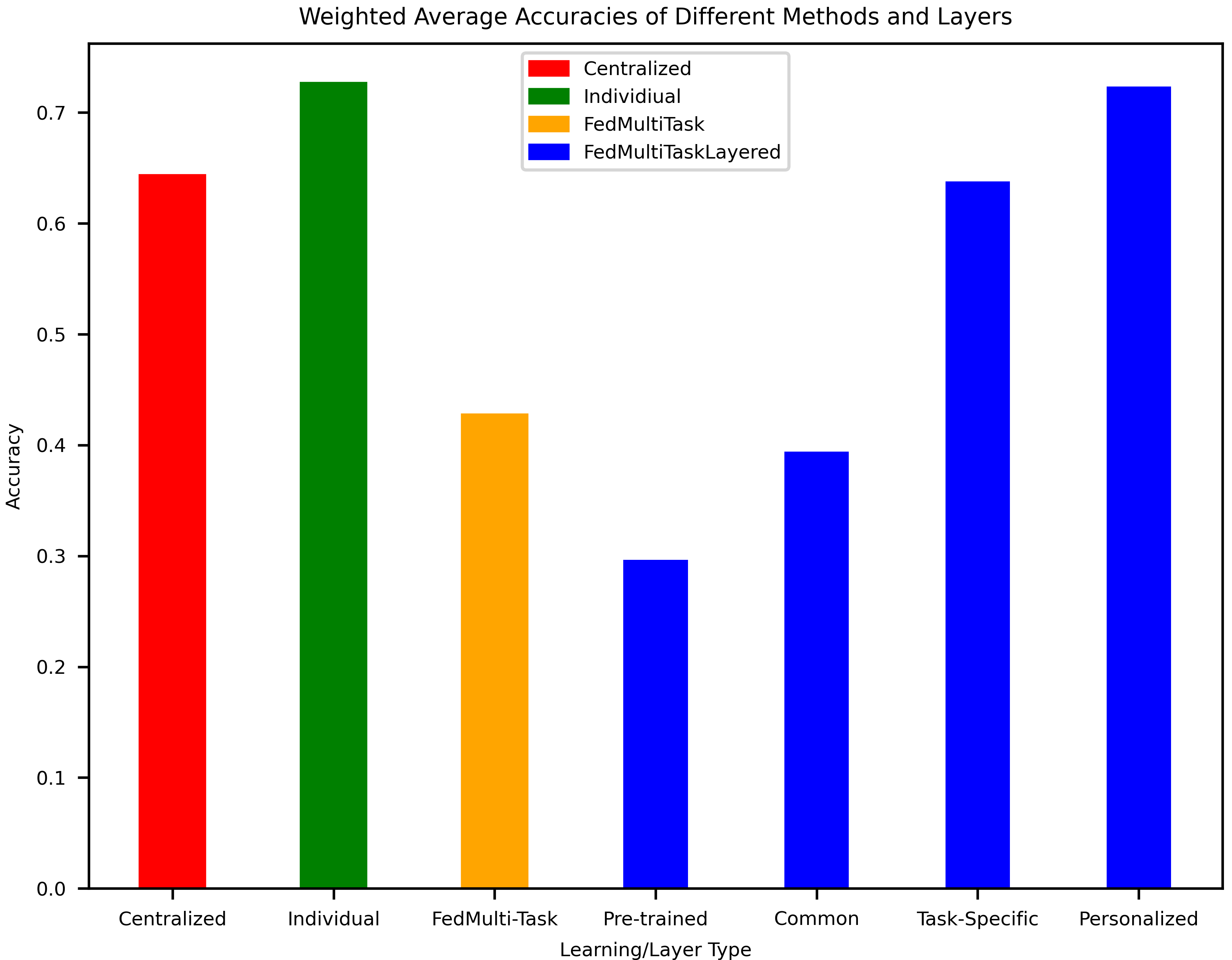}
	\caption{Weighted Average of Accuracies of 13 Clients For Different Methods/Layers.}
\vspace*{-0.4cm}
	\label{fig:multi_task_layers}

\end{figure}

The performance of training with different layers can be seen in Figure~\ref{fig:multi_task_layers}. The percentages are the weighted average accuracies of all the clients. The first three columns present the baselines, centralized, individual (one-task) and simple federated learning (multi-task). The latest personalized layer has an average weighted accuracy of 72.4\%, whereas the individual learning's accuracy is 72.6\% and federated multi-task is 42.7\%. From these results, we can say that personalization works quite well, similar to training the clients individually, and the accuracy loss in task-specific layers is tolerable, but the pre-trained and common layers' accuracies are not close to this individual training baseline. Also, the layered hierarchy achieves more success than simple federated multi-task training. Federated multi-task transfer learning achieved similar results to individual training and better results than centralized training. One may argue that each client can be trained with its own data. However, collecting labelled data to train a model is a difficult process, and by using a transfer learning approach, the clients can get similar results with a limited amount of labelled data, which is used in the personalization step. Due to space limitations, we cannot present the accuracy results for each client. However, we observe that at every layer, the accuracy increases and gets closer to the individual training baseline. 



\section{Conclusion \& Future Studies }
\label{sec:Conclusion}

This paper focuses on the comparison between various models of centralized learning, federated learning and federated multi-task transfer learning combined with the DeepConvLSTM based architecture for human activity classification and device position identification with motion sensor data. Models are trained using the integrated OpenHAR dataset containing ten smaller datasets. In multi-task federated transfer learning, the obtained success rates are similar to the baseline's success (individual training) and are better than a fully centralized approach. For future studies, different averaging algorithms than FedAvg and different classifiers' effects on success will be analyzed. In addition, experiments can be implemented with different percentages of training data to analyze the effect of the data amount.

\vspace*{-0.25cm}
\small 
\bibliography{output}

\begin{thebibliography}{11}
\providecommand{\natexlab}[1]{#1}
\providecommand{\url}[1]{\texttt{#1}}
\expandafter\ifx\csname urlstyle\endcsname\relax
  \providecommand{\doi}[1]{doi: #1}\else
  \providecommand{\doi}{doi: \begingroup \urlstyle{rm}\Url}\fi

\bibitem[Beutel et~al.(2020)Beutel, Topal, Mathur, Qiu, Fernandez-Marques, Gao, Sani, Li, Parcollet, de~Gusm{\~a}o, et~al.]{Flower}
Daniel~J Beutel, Taner Topal, Akhil Mathur, Xinchi Qiu, Javier Fernandez-Marques, Yan Gao, Lorenzo Sani, Kwing~Hei Li, Titouan Parcollet, Pedro Porto~Buarque de~Gusm{\~a}o, et~al.
\newblock Flower: A friendly federated learning research framework.
\newblock \emph{arXiv preprint arXiv:2007.14390}, 2020.

\bibitem[Chen et~al.(2020)Chen, Qin, Wang, Yu, and Gao]{Fedhealth}
Yiqiang Chen, Xin Qin, Jindong Wang, Chaohui Yu, and Wen Gao.
\newblock Fedhealth: A federated transfer learning framework for wearable healthcare.
\newblock \emph{IEEE Intelligent Systems}, 35\penalty0 (4):\penalty0 83--93, 2020.

\bibitem[Ek et~al.(2020)Ek, Portet, Lalanda, and Vega]{portet20}
Sannara Ek, Fran\c{c}ois Portet, Philippe Lalanda, and German Vega.
\newblock Evaluation of federated learning aggregation algorithms: Application to human activity recognition.
\newblock In \emph{Adjunct Proceedings UbiComp/ISWC '20}, page 638–643, 2020.

\bibitem[Garcia-Gonzalez et~al.(2020)Garcia-Gonzalez, Rivero, Fernandez-Blanco, and Luaces]{anguita2013public}
Daniel Garcia-Gonzalez, Daniel Rivero, Enrique Fernandez-Blanco, and Miguel~R. Luaces.
\newblock A public domain dataset for real-life human activity recognition using smartphone sensors.
\newblock \emph{Sensors}, 20\penalty0 (8), 2020.

\bibitem[Ke{\c{c}}eci et~al.(2022)Ke{\c{c}}eci, Shaqfeh, Mbayed, and Serpedin]{Multi-Task}
Cihat Ke{\c{c}}eci, Mohammad Shaqfeh, Hayat Mbayed, and Erchin Serpedin.
\newblock Multi-task and transfer learning for federated learning applications.
\newblock \emph{arXiv preprint arXiv:2207.08147}, 2022.

\bibitem[McMahan et~al.(2017)McMahan, Moore, Ramage, Hampson, and y~Arcas]{FedAvg}
Brendan McMahan, Eider Moore, Daniel Ramage, Seth Hampson, and Blaise~Aguera y~Arcas.
\newblock Communication-efficient learning of deep networks from decentralized data.
\newblock In \emph{Artificial Intelligence and Statistics}, pages 1273--1282. PMLR, 2017.

\bibitem[Ordóñez and Roggen(2016)]{DCL}
Francisco~Javier Ordóñez and Daniel Roggen.
\newblock Deep convolutional and lstm recurrent neural networks for multimodal wearable activity recognition.
\newblock \emph{Sensors}, 16\penalty0 (1), 2016.

\bibitem[Presotto et~al.(2022{\natexlab{a}})Presotto, Civitarese, and Bettini]{presotto}
Riccardo Presotto, Gabriele Civitarese, and Claudio Bettini.
\newblock Semi-supervised and personalized federated activity recognition based on active learning and label propagation.
\newblock \emph{Personal and Ubiquitous Computing}, 26:\penalty0 1--18, 06 2022{\natexlab{a}}.

\bibitem[Presotto et~al.(2022{\natexlab{b}})Presotto, Civitarese, and Bettini]{presottofedclar}
Riccardo Presotto, Gabriele Civitarese, and Claudio Bettini.
\newblock Fedclar: Federated clustering for personalized sensor-based human activity recognition.
\newblock In \emph{2022 IEEE International Conference on Pervasive Computing and Communications (PerCom)}, pages 227--236, 2022{\natexlab{b}}.

\bibitem[Siirtola et~al.(2018)Siirtola, Koskim\"{a}ki, and R\"{o}ning]{OpenHAR}
Pekka Siirtola, Heli Koskim\"{a}ki, and Juha R\"{o}ning.
\newblock Openhar: A matlab toolbox for easy access to publicly open human activity data sets.
\newblock In \emph{Adjunct Proceedings UbiComp/ISWC '18}, page 1396–1403, New York, NY, USA, 2018.

\bibitem[Sztyler and Stuckenschmidt(2016)]{sztyler2016body}
Timo Sztyler and Heiner Stuckenschmidt.
\newblock On-body localization of wearable devices: An investigation of position-aware activity recognition.
\newblock In \emph{2016 IEEE International Conference on Pervasive Computing and Communications (PerCom)}, pages 1--9, 2016.

\end{thebibliography}

\end{document}